\documentclass[a4paper,fleqn]{cas-sc}

\usepackage[authoryear]{natbib}

\usepackage{amsthm,amsmath,amssymb,mathrsfs}
\usepackage{pifont}
\usepackage{algorithm}
\usepackage{algorithmic}
\usepackage{graphicx}
\usepackage{subfigure}
\usepackage{threeparttable}
\usepackage{color}
\usepackage{booktabs}
\usepackage{multirow}
\usepackage{array}
\usepackage{soul, color, xcolor}

\usepackage{enumerate}
\usepackage{rotating}
\usepackage{adjustbox}

\usepackage{etoolbox}

\makeatletter
\renewcommand{\fnum@figure}{Fig. \thefigure.\@gobble }
\makeatother

\def\tsc#1{\csdef{#1}{\textsc{\lowercase{#1}}\xspace}}
\tsc{WGM}
\tsc{QE}
\tsc{EP}
\tsc{PMS}
\tsc{BEC}
\tsc{DE}

\begin{document}
\let\WriteBookmarks\relax
\def\floatpagepagefraction{1}
\def\textpagefraction{.001}
\let\printorcid\relax

\shorttitle{}


\title [mode = title]{How defensive driving enhances driving safety: A driving simulator study on drivers' defensive driving behaviors}      

\author[1]{\textcolor{black}{Xinzheng Wu}}[]
\credit{Conceptualization, Formal analysis, Methodology, Visualization, Writing - original draft}
\affiliation[1]{organization={College of Automotive and Energy Engineering, Tongji University},
    addressline={No. 4800 Cao'an Road.}, 
    city={Shanghai},
    postcode={201804}, 
    country={China}}

\author[1]{\textcolor{black}{Junyi Chen}}[]
\credit{Formal analysis, Project administration, Validation, Writing - review \& editing}
\cormark[1]
\ead{chenjunyi@tongji.edu.cn}

\author[1]{\textcolor{black}{Shaolingfeng Ye}}[]
\credit{Data curation, Software, Validation, Writing - review \& editing}

\author[1]{\textcolor{black}{Yong Shen}}[]
\credit{Investigation, Resources, Supervision, Writing - review \& editing}

\cortext[cor1]{Corresponding author}

\begin{abstract}
Defensive driving is widely recognized as an advanced driving skill. However, whether and how defensive driving affects driving safety remains insufficiently investigated. This study examines the behavioral characteristics of defensive driving, its impact on driving safety, and the underlying mechanisms.
First, defensive driving is defined regarding operational timing and application scenario. Then, 82 participants are recruited for driving simulator experiments, with their behavioral and eye movement data being collected. 
Following the experiments, participants are categorized into groups based on the frequency of defensive driving behaviors exhibited. Finally, both inter-group and inter-trial comparisons are performed on the experimental data.
Experimental results demonstrate that in the inter-group comparison, the high defensive driving capability group exhibits higher acceleration and deceleration magnitudes, lower average speeds, and larger average absolute yaw angles compared to the low capability group, alongside shorter fixation durations and reduced fixation frequencies. Moreover, we observe that these participants tend to initiate defensive or evasive actions earlier, resulting in lower scenario risk. Regarding the inter-trial comparison, we observe similar trends exclusively in the low capability group, whereas most metrics show no significant differences between Trial 1 and Trial 2 in the high capability group.
These results reveal that drivers possessing defensive driving capabilities tend to execute more intense driving maneuvers and identify risks and take action earlier, thereby enhancing driving safety.
Findings support the promotion of defensive driving and provide a basis for relevant training programs. Meanwhile, they offer insights for the training of autonomous driving algorithms with defensive driving capabilities.

\end{abstract}


\maketitle

\section{Introduction} \label{sec1}
According to the Road Safety Annual Report 2025 published by the International Transport Forum, nearly 80,000 traffic fatalities occurred across 35 countries worldwide in 2024 \citep{ITF2025}. Meanwhile, data from the UK's Department for Transport indicates that 52\% of fatal collisions resulted from driver behavior or inexperience \citep{GovUK2024RoadCasualties}, highlighting the critical need to improve drivers' driving skills to boost road safety \citep{ratchaneepun2026transferability}. Furthermore, reports from China's National Bureau of Statistics \citep{StatsGovCN2024TrafficAccidents} and the U.S. National Highway Traffic Safety Administration (NHTSA) \citep{national2026overview} both reveal a rising trend in traffic accidents involving cyclists and pedestrians, who are regarded as road users with high uncertainty. This emphasizes the necessity for drivers to possess the capability to drive proactively in high-uncertainty scenarios.

Defensive driving, as an advanced driving technique, is considered promising for enhancing road safety and reducing traffic accidents, and is recommended by numerous organizations. For instance, the New York State Driver's Manual recommends defensive driving behaviors such as 'Be prepared and look ahead', 'Maintain the correct speed', and 'Signal before you turn or change lanes' \citep{DMVNY2020DefensiveDriving}. Additionally, the National Safety Council (NSC) offers comprehensive defensive driving courses for both the general public and professional drivers \citep{NSC2026DefensiveDrivingCourses}. Concurrently, defensive driving has garnered widespread attention within academia. \cite{stahl2014anticipation} proposed anticipatory driving, defining it as a high-level cognitive competence based on environmental cue recognition, which can be regarded as a subset of defensive driving. However, their simulator experiments demonstrated that the corresponding driving behaviors improved safety only under certain conditions. Based on core defensive driving principles such as maintaining a safe distance from the lead vehicle and avoiding blind spots, \cite{bhavsar2025evaluating} developed an assessment method for defensive driving behaviors based on naturalistic driving data. Furthermore, \cite{zhang2026intelligent} constructed an intelligent defensive driving framework for autonomous vehicles based on human defensive driving experience and validated its effectiveness in enhancing driving safety. The aforementioned studies all emphasize the core principle of defensive driving, namely the identification and mitigation of potential risks. However, there is a lack of systematic and formal definitions regarding the operational timing and application scenarios of defensive driving. Furthermore, despite extensive research and recommendations, whether and how defensive driving influences driving safety remains an open question.

\begin{figure}[pos=t] 
      \centering
      \includegraphics[width=\linewidth]{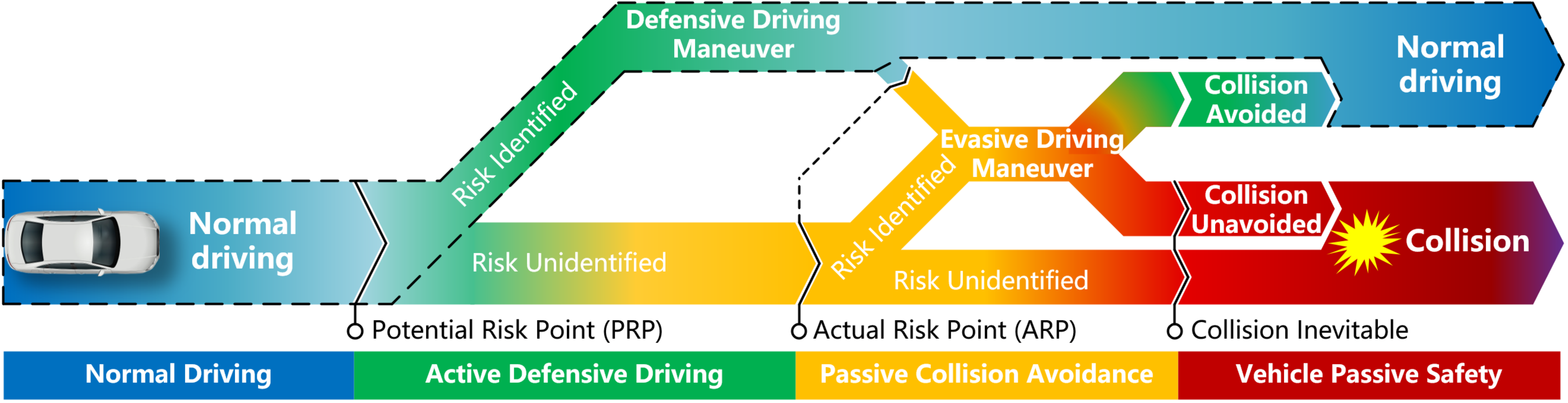}
      \caption{The definition of potential risk point (PRP) and actual risk point (ARP), with a successful defensive driving path circled in dashed lines.}
      \label{prp_arp}
\end{figure}

To address this issue, we first define two key temporal points, the Potential Risk Point (PRP) and the Actual Risk Point (ARP), representing the moments at which potential and actual risks emerge, as illustrated in Fig. \ref{prp_arp}. Specifically, potential risk refers to probabilistic threats arising from the uncertainty of other traffic participants, even if a driver does not respond to such risks, a collision will not occur immediately in the short term. In contrast, actual risk denotes deterministic threats resulting from explicit trajectory conflicts between the ego vehicle and other traffic participants, where a collision becomes inevitable if evasive action is not taken. Based on the aforementioned definitions, we define defensive driving as \textit{a high-level driving behavior characterized by the accurate identification of the potential risk point (PRP) and the timely execution of appropriate proactive maneuvers to mitigate potential risks, thereby restoring the vehicle to a normal driving state and avoiding exposure to collision risks}. 

It is noteworthy that, as illustrated in Fig. \ref{prp_arp}, vehicles may still encounter actual risks even after executing a defensive driving maneuver (e.g., when approaching a blind spot, a pedestrian or cyclist may still rush out unexpectedly despite prior deceleration). Consequently, when the ARP emerges, drivers still need to perform evasive driving maneuvers. Furthermore, even in the absence of defensive driving maneuvers, drivers might still achieve successful collision avoidance and restore normal driving states solely through timely and effective evasive driving maneuvers once the ARP appears. Although the safety outcome in such scenarios is consistent with that of defensive driving, the process inevitably involves aggressive behaviors such as emergency braking or sharp steering. This significantly compromises driving efficiency and comfort. Moreover, in the context of autonomous vehicles, it severely undermines passenger trust \citep{kim2025disuse, cai2026research}.

Based on the definition of defensive driving, this study carefully designs experimental scenarios (see Section \ref{scenario}) and conducts driving simulator experiments to investigate whether and how defensive driving affects driving safety. In previous studies, driving simulators have been widely adopted for collecting driver behavior data \citep{vaezipour2018simulator,zhang2023driving}, eye movement information \citep{he2021anticipatorya, gideon2025simulator}, and subjective ratings \citep{kuhn2024let, wu2025riseea}, owing to their low cost, high safety, and reproducibility. In this study, driver behavioral data (such as braking and steering) can serve as valid indicators for determining whether a driver has executed a defensive driving maneuver. Concurrently, eye movement data can not only be utilized to assess driver states such as hazard perception \citep{qin2025eye}, distraction \citep{qiao2025driver, kummerer2026mitigating}, and fatigue \citep{zhang2024driving}, but also be regarded as a distinct driving behavior pattern (e.g., fixation and saccade behaviors) \citep{zhao2025comparing, denk2025influence} to investigate the characteristics of defensive driving.

In summary, although defensive driving is widely recognized as a key safe driving strategy, its impact on driving safety and the underlying mechanisms remain insufficiently investigated. To address these gaps, this study conducts a driving simulator experiment to collect both behavioral (e.g., braking, steering) and eye movement data and performs subsequent analyses to address the following research questions:

\begin{enumerate}[1)]
    \item RQ1: What are the behavioral characteristics of defensive driving?
    \item RQ2: What is the impact of defensive driving on driving safety?
    \item RQ3: Through what mechanisms does defensive driving influence driving safety?
    \item RQ4: How can defensive driving capability be acquired?
\end{enumerate}

\section{Materials and Method} \label{sec2}

\subsection{Participants}
A total of 82 licensed human drivers were recruited in this study. Participants were primarily recruited via social media platforms within the university community, including students, security personnel, and other university staff. Screening was based on demographic information, driving background, and self-reported driving experience collected through the recruitment questionnaire. The demographic information of these participants is summarized in Table \ref{tab1}. As shown in Table \ref{tab1}, the recruited drivers cover a wide range of ages (mean 24.17, SD 3.88), driving years (mean 3.75, SD 2.68), driving frequencies, and self-rated driving abilities. This diversity helps ensure the collection of driving data across various levels of defensive driving capability, enabling an investigation into the impact of this capability on driving behavior and safety.

\begin{table}[pos=b]
\renewcommand{\arraystretch}{1.2}
\caption{Information of the participants.}
\label{tab1}
\begin{center}
\begin{tabular}{ m{0.17\linewidth} m{0.35\linewidth}  m{0.12\linewidth}<{\centering} m{0.12\linewidth}<{\centering} }
\toprule
\textbf{Information} & \textbf{Characteristics} & \textbf{Number} & \textbf{Ratio}  \\
\midrule
\multirow{2}{*}{Gender} & Male  &  67 & 81.71\% \\
~ & Female & 15 & 18.29\% \\ \hline
\multirow{3}{*}{Age} & 20-25  &  73 & 89.02\% \\
~ & 26-30 & 6 & 7.32\% \\
~ & >30 & 3 & 3.66\% \\ \hline
\multirow{4}{.8\linewidth}{Driving Years} & $\le$2  &  29 & 35.37\% \\
~ & 3-5 & 42 & 51.22\% \\
~ & 6-10 & 10 & 12.20\% \\
~ & >10 & 1 & 1.22\% \\ \hline
\multirow{4}{.8\linewidth}{Driving Frequency} & Less than once a month & 37 & 45.12\% \\
~ & At least once a month & 30 & 36.59\% \\
~ & At least once a week & 11 & 13.41\% \\ 
~ & At least once a day & 4 & 4.88\% \\ \hline
\multirow{4}{\linewidth}{Self-assessment of Driving Ability} & Novice & 14 & 17.07\% \\
~ & Intermediate & 24 & 29.27\% \\
~ & Proficient & 32 & 39.02\% \\ 
~ & Expert & 12 & 14.63\% \\
\bottomrule
\end{tabular}
\end{center}
\end{table}

\subsection{Experimental scenarios} \label{scenario}

As illustrated in Fig. \ref{prp_arp}, the key trigger for implementing defensive driving maneuvers lies in the presence of potential risks. Therefore, it is necessary to design scenarios containing such risks to capture human driver behavior. As discussed previously, potential risks stem from the uncertainty of other traffic participants. This study further categorizes this uncertainty into two dimensions: existence uncertainty and motion uncertainty. 

Specifically, existence uncertainty is primarily induced by physical blind spots. Occlusion caused by road infrastructure or obstacles prevents drivers from perceiving the status within these areas, making it impossible to determine whether other traffic participants are about to emerge. Based on the cause of formation, existence uncertainty-related scenarios can be further categorized into static blind spot scenarios (caused by stationary obstructions) and dynamic blind spot scenarios (caused by moving obstructions). 

Conversely, motion uncertainty arises from irregular behaviors exhibited by visible traffic participants. Even when drivers can observe other traffic participants, their potentially abnormal driving behaviors (e.g., aggressive takeovers or sudden lane changes) may still result in abrupt trajectory alterations. Depending on whether relevant cues exist to anticipate these irregular behaviors, the scenarios can be categorized into predictable motion uncertainty scenarios (also referred to as anticipatory scenarios in \cite{he2021anticipatorya}) and unpredictable motion uncertainty scenarios.

The experimental scenarios designed for this study are presented in Table \ref{tab2}. This table provides detailed descriptions of each scenario, along with specific definitions of the PRP and ARP. In particular, in the two scenarios involving cyclists, Time Headway (THW) is used as the trigger metric for cyclist initiation. Consequently, since each participant drives at a different ego velocity ($v_{Ego}$), the actual triggering distance ($d_{trigger}$) varied across participants. Moreover, apart from the specific surrounding vehicle (SV) in direct interaction with the ego vehicle, the remaining SVs serve to restrict the available drivable area of the ego vehicle, thus inducing the intended conflicts. It should be noted that, for the unpredictable motion uncertainty scenario, due to the absence of any cues for drivers to identify the PRP, even experienced drivers struggle to execute defensive driving maneuvers in advance. Consequently, such scenarios fall outside the scope of this study.

\begin{sidewaystable*}[tp]
\renewcommand{\arraystretch}{1.3}
\centering
\caption{Illustrations, descriptions, and specific definitions of PRP and ARP for each experimental scenario.}
\begin{tabular}{
  m{0.08\linewidth}
  >{\raggedright}m{0.08\linewidth}
  p{0.4\linewidth}<{\centering}
  m{0.2\linewidth}
  m{0.07\linewidth}
  m{0.07\linewidth}
}
\toprule
\textbf{Uncertainty} & \textbf{Category} & \textbf{Scenario illustration} & \textbf{Scenario description} & \textbf{PRP} & \textbf{ARP} \\
\midrule
\multirow{7}{0.08\linewidth}{\centering Existence uncertainty} 
& Static blind spot scenario & 
\adjustbox{valign=c}{\includegraphics[width=\linewidth]{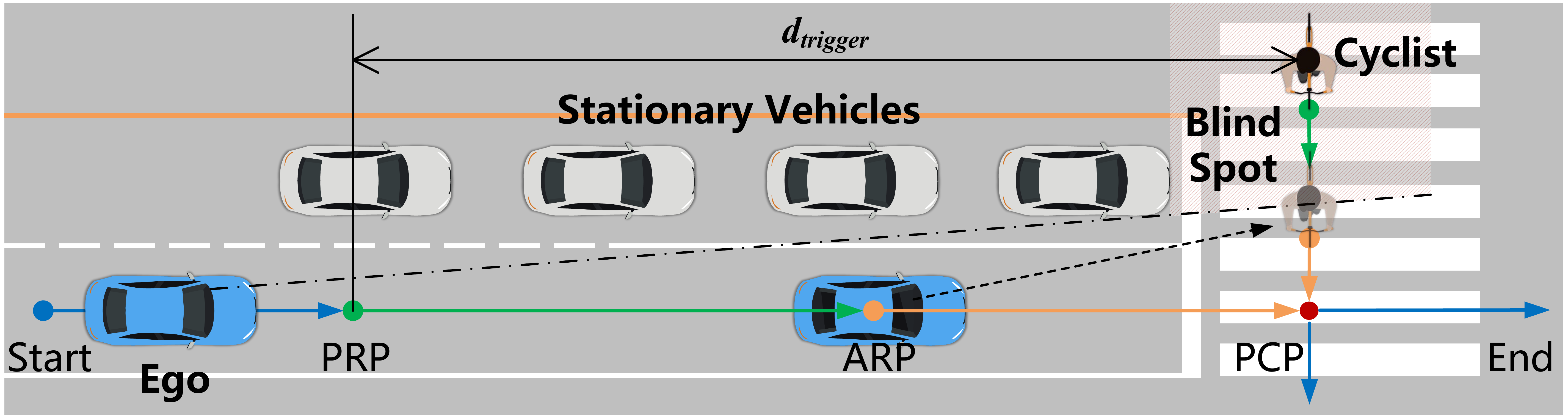}}
& 
The ego vehicle travels in the right lane, with a cyclist occluded by a row of stationary vehicles in the left lane at the intersection. The cyclist begins moving at $4 m/s$ when the ego vehicle's THW to the Potential Collision Point (PCP) $THW = \frac{d_{trigger}}{v_{Ego}} < 4.4s$. & 
Cyclist starts moving & 
Ego detects cyclist \\
\cline{2-6}
& Dynamic blind spot scenario & 
\adjustbox{valign=c}{\includegraphics[width=\linewidth]{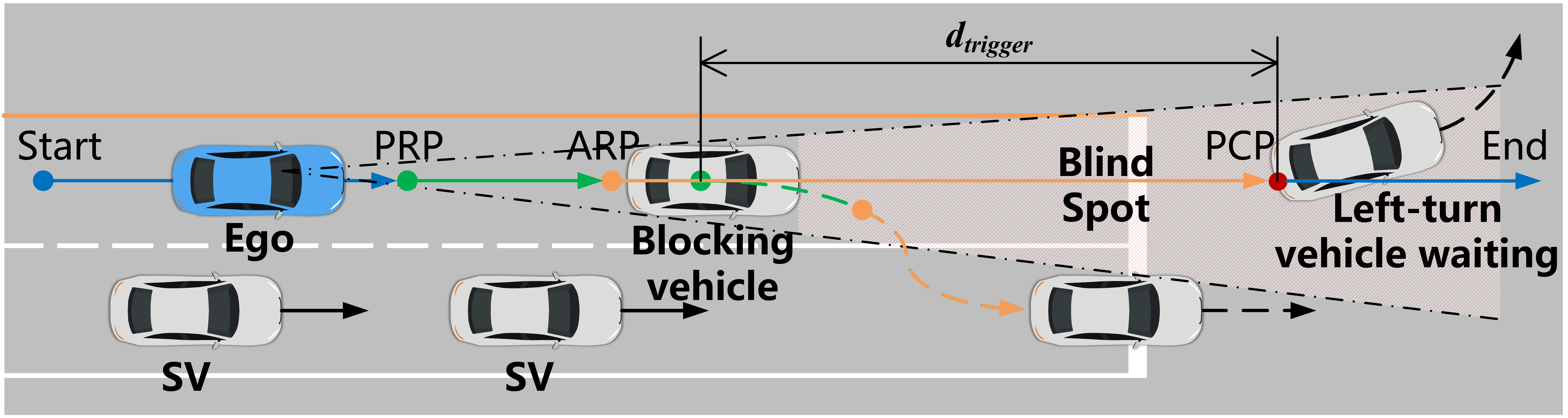}}
& 
The ego vehicle travels in the left lane, with a blocking vehicle (driving at $14m/s$) obstructing a stationary vehicle preparing to turn left at the upcoming intersection. When $d_{trigger} < 25 m$, the blocking vehicle moves to the right, revealing the hidden vehicle.& 
Blocking vehicle initiates lane change & 
Ego detects left-turn vehicle \\
\hline
Motion uncertainty& Predictable motion uncertainty scenario & 
\adjustbox{valign=c}{\includegraphics[width=\linewidth]{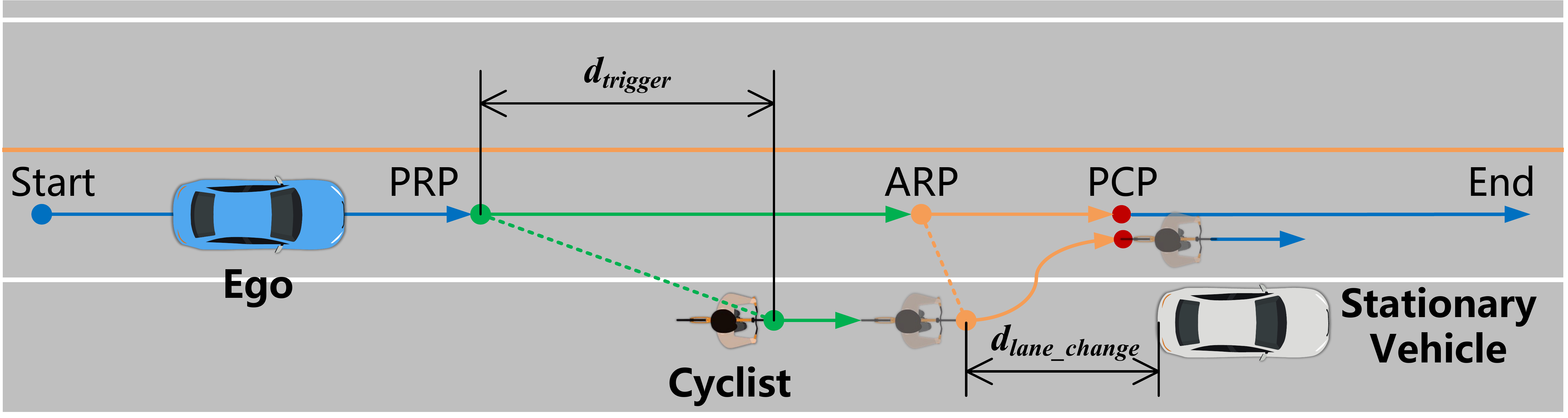}}
& 
The ego vehicle travels in a single lane, with a cyclist riding ahead on the adjacent non-motorized lane, whose path is blocked by a stationary vehicle. The cyclist begins moving at $3 m/s$ when $THW = \frac{d_{trigger}}{v_{Ego}} < 6.0 s$. Subsequently, when $d_{lane\_chang}<15m$, the cyclist turns left to avoid the stationary vehicle and returns to the original lane after overtaking it. & 
Cyclist starts moving & 
Cyclist initiates lane change \\
\bottomrule
\end{tabular}
\label{tab2}
\end{sidewaystable*}

\subsection{Experimental equipment and procedure}
\begin{figure}[pos=b] 
      \centering
      \includegraphics[width=\linewidth]{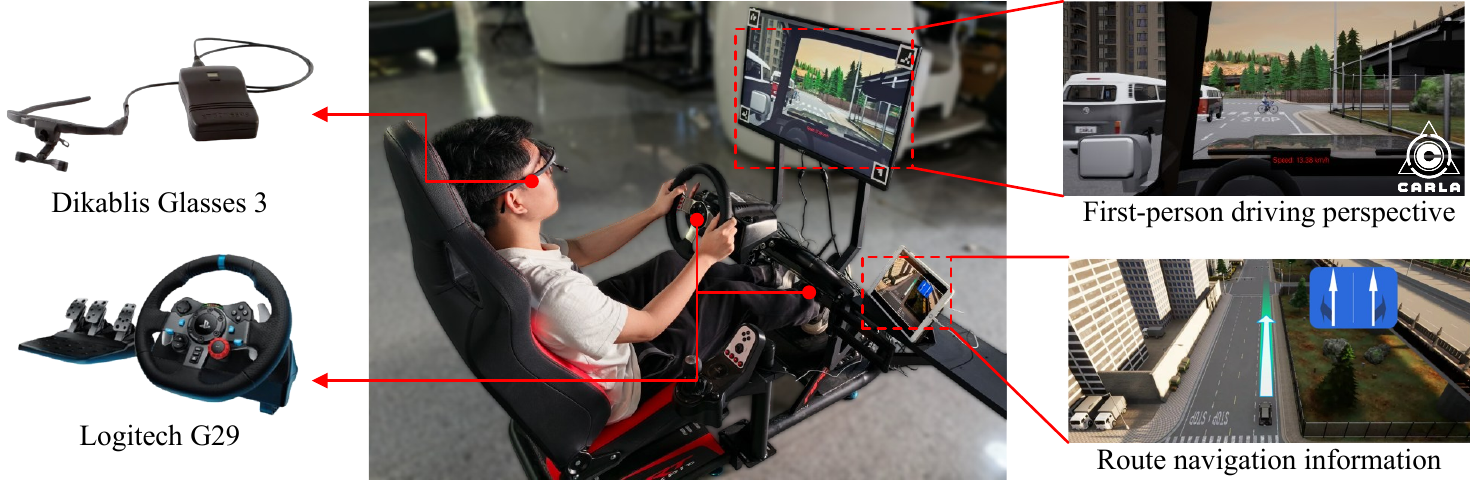}
      \caption{Illustration of the experimental setup.}
      \label{driving_simulator}
\end{figure}

The experimental setup is illustrated in Fig. \ref{driving_simulator}. The experiment is conducted on the CARLA simulation platform \citep{dosovitskiy2017carla}, where a high-resolution monitor displays the first-person driving perspective generated by the simulator. Participants operate the ego vehicle using a Logitech G29 steering wheel and pedal set. Throughout the experiment, a Dikablis Glasses 3 head-mounted eye tracker is used to continuously record the participants' eye movements. Notably, a route navigation map is displayed in the front-right field of view to guide participants along the correct route, ensuring the occurrence of conflicts with the intended interactive objects.

Before the experiment, all participants are fully informed of the potential risks and discomfort, privacy protection measures, and their right to withdraw at any time. To avoid creating expectancy effects regarding defensive driving, participants are only given a vague description of the study's purpose (i.e., "driving behavior data collection") and are deliberately kept unaware of the specific focus on defensive driving or any scenario details. This ensures the acquisition of natural behavioral responses. After signing the informed consent form, participants are first fitted with the eye tracker and undergo a calibration procedure. Then, they are required to complete approximately 2 minutes of free driving in the simulator to adapt to the experimental environment and equipment.

At the beginning of the formal experiment, the ego vehicle is initialized at zero speed at a considerable distance away from the Potential Conflict Point (PCP) (as illustrated in the scenario diagrams in Table \ref{tab2}). Participants are instructed to initiate the ego vehicle and adjust its speed freely based on their own driving experience. Each participant drives all the three scenarios listed in Table \ref{tab2} in a randomized order, completing two repetitions per scenario. The entire experiment lasts approximately 45 minutes.



\subsection{Data analysis method} \label{sec_data_analysis}

After the completion of the experiment, a total of 492 samples are obtained (82 participants $\times$ 3 scenarios $\times$ 2 trials). Given that participants determine the ego vehicle's speed on their own, the PRP and ARP vary across samples. Therefore, we first calculate the specific PRP and ARP for each driving sample based on the definitions in Table \ref{tab2}. Subsequently, kinematic and eye movement metrics are extracted for two distinct time intervals: from PRP to ARP, and from PRP to scenario end. Furthermore, risk indicators, behavioral indicators, and binary indicators are computed. All calculated statistics are summarized in Table \ref{tab3}.

For kinematic metrics, we focus on both longitudinal parameters (e.g. speed, acceleration, and deceleration) and lateral parameters (e.g. vehicle yaw angle and lateral displacement) to facilitate the identification of the motion characteristics of defensive driving. For eye movement metrics, two categories of metrics are extracted: 1) pupil-related metrics, which reflect the driver's perception of and response to risk or hazards, and 2) fixation/saccade-related metrics, which elucidate the visual observation patterns inherent in defensive driving.

In addition to the aforementioned directly measurable metrics, we further compute additional indicators based on the collected trajectories to assess driving risk, behavioral characteristics, and task efficiency. It is noteworthy that driving performance during the defensive driving phase (from PRP to ARP) inevitably affects overall scenario safety and efficiency. Consequently, all subsequent indicators are calculated over the time interval spanning from the PRP to the end of the scenario. The calculated indicators are detailed as follows.

\begin{itemize}
    \item \textbf{Minimum Time To Collision (min TTC)}: TTC is a widely adopted risk indicator that estimates the time remaining before a collision occurs between the ego vehicle and a leading obstacle \citep{vogel2003comparison}. In this study, the minimum TTC observed is used to characterize the most hazardous situation, serving as a measure of driving safety.
    \item \textbf{Maximum Discretized Normalized Drivable Area (max DNDA)}: Unlike TTC which quantifies solely longitudinal risk, DNDA characterizes risk in both longitudinal and lateral dimensions based on the concept of the Drivable Area \citep{wu2022risk}. The DNDA value ranges from 0 (no SV within the drivable area) to 1 (collision). In this study, the maximum DNDA observed is utilized as the risk indicator.
    \item \textbf{Post Encroachment Time (PET)}: PET \citep{johnsson2018search} is a post-hoc risk indicator that measures the time gap between two traffic participants arriving at the PCP, but only if their trajectories actually conflict. Therefore, PET is unavailable when the vehicle performs a lateral defensive/evasive maneuver that effectively avoids a trajectory conflict.
    \item \textbf{Distance at Action (DistAct)}: DistAct quantifies the distance between the ego vehicle and the PCP at the moment the driver initiates the first defensive or evasive maneuver.
    \item \textbf{Task Completion Time (TCT)}: TCT is defined as the duration from the experiment's start to the moment the ego vehicle reaches the predefined scenario endpoint. Therefore, TCT is unavailable when the ego vehicle fails to reach this endpoint.
    \item \textbf{Collision} and \textbf{Task Completion (TC)}: Two binary indicators that represent whether a collision occurrs and whether the driving task is successfully completed. Specifically, in this study, task completion is evaluated exclusively in the absence of collisions. Therefore, a 'False' TC outcome indicates that the driver executes an excessively abrupt braking maneuver, resulting in a complete stop before reaching the predefined scenario endpoint
\end{itemize}

To address the four research questions posed in Section \ref{sec1}, the data analysis is conducted from two perspectives: 1) comparing data differences between groups with high and low defensive driving capabilities, and 2) comparing data differences within the same group of participants between the first and second experimental trials. A summary of the specific data analysis methods is also provided in Table \ref{tab3}. For inter-group comparison, since the data does not follow a normal distribution, the Mann-Whitney U test is used for continuous variables to conduct non-parametric analysis, whereas the chi-square test is employed for binary variables. In contrast, for inter-trial comparison, because the data from Trial 1 and Trial 2 can be regarded as paired samples, the Wilcoxon signed‑rank test is applied for continuous variables, and McNemar’s test is utilized for binary variables.

\begin{table}[pos=t]
\renewcommand{\arraystretch}{1.2}
\caption{Summary of metrics/indicators and corresponding data analysis methods.}
\label{tab3}
\begin{center}
\begin{tabular}{ m{0.08\linewidth} m{0.36\linewidth} m{0.03\linewidth} m{0.21\linewidth}<{\centering}  m{0.2\linewidth}<{\centering}} 
\toprule
\multirow{2}{\linewidth}{\textbf{Types}} & \multirow{2}{\linewidth}{\textbf{Name}} & \multirow{2}{\linewidth}{\textbf{Unit}} & \multicolumn{2}{c}{\textbf{Data Analysis Methods}}  \\ \cline{4-5}
& & & \textbf{ Inter-group Comparison} & \textbf{ Inter-trial Comparison} \\
\midrule
\multirow{10}{\linewidth}{Kinematic Metrics} 
& Maximum longitudinal deceleration (max Dec)       & $m/s^2$ & \multirow{10}{*}{Mann-Whitney U Test} & \multirow{10}{*}{Wilcoxon Signed-rank Test} \\ \cline{2-3}
& Maximum longitudinal acceleration (max Acc)       & $m/s^2$ & & \\ \cline{2-3}
& Mean longitudinal speed (mean Speed)              & $m/s$   & & \\ \cline{2-3}
& Maximum longitudinal speed (max Speed)            & $m/s$   & & \\ \cline{2-3}
& Mean absolute yaw angle (mean Yaw)                & $deg$   & & \\ \cline{2-3}
& Standard deviation of absolute yaw angle (SD Yaw) & $deg$   & & \\ \cline{2-3}
& Maximum absolute yaw angle (max Yaw)          & $deg$   & & \\ \cline{2-3}
& Maximum lateral displacement (max LD)             & $m$     & & \\ \cline{2-3}
& Standard deviation of lateral displacement (SDLD) & $m$ & & \\  \hline

\multirow{10}{\linewidth}{Eye Movement Metrics} 
& Average Pupil Diameter (APD)           & $pixel$ & \multirow{10}{*}{Mann-Whitney U Test} & \multirow{10}{*}{Wilcoxon Signed-rank Test} \\ \cline{2-3}
& Maximum Pupil Diameter (MPD)           & $pixel$ & & \\ \cline{2-3}
& Average Fixation Duration (AFD)        & $ms$    & & \\ \cline{2-3}
& Average Saccade Duration (ASD)         & $ms$    & & \\ \cline{2-3}
& Maximum Fixation Duration (MFD)        & $ms$    & & \\ \cline{2-3}
& Maximum Saccade Duration (MSD)         & $ms$    & & \\ \cline{2-3}
& Maximum Saccade Angle (MSA)            & $deg$   & & \\ \cline{2-3}
& Standard Deviation of Saccade Angle (SAS) & $deg$    & & \\ \cline{2-3}
& Fixation Frequency (FF)                & $Hz$    & & \\ \cline{2-3}
& Saccade Frequency (SF)                 & $Hz$    & & \\ \hline

\multirow{4}{\linewidth}{Risk Indicators} 
& Minimum Time to Collision (min TTC)    & $s$     & \multirow{4}{*}{Mann-Whitney U Test} & \multirow{4}{*}{Wilcoxon Signed-rank Test} \\ \cline{2-3}
& Maximum Discretized Normalized Drivable Area (max DNDA) & / & & \\ \cline{2-3}
& Post Encroachment Time (PET)           & $s$     & & \\ \hline

\multirow{2}{\linewidth}{Behavioral Indicators} 
& Distance at Action (DistAct)                  & $m$     & \multirow{2}{*}{Mann-Whitney U Test} & \multirow{2}{*}{Wilcoxon Signed-rank Test} \\ \cline{2-3}
& Task Completion Time (TCT)                  & $s$     & & \\ \hline

\multirow{2}{\linewidth}{Binary Indicators} 
& Collision                              & /       & \multirow{2}{*}{Chi-square Test} & \multirow{2}{*}{McNemar's Test} \\ \cline{2-3}
& Task completion (TC)                       & /       & & \\
\bottomrule
\end{tabular}
\end{center}
\end{table}

\section{Results}

\subsection{Data screening}

\begin{figure}[pos = b] 
      \centering
      \includegraphics[width=.4\linewidth]{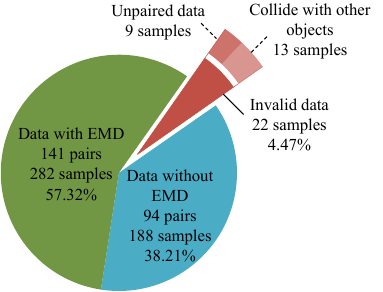}
      \caption{The distribution of the screened data.}
      \label{data_screening}
\end{figure}

The distribution of the screened data is shown in Fig. \ref{data_screening}. From the initial 492 samples, 22 are identified as invalid. Specifically, 13 samples result in collisions prior to reaching the expected interaction zone (i.e., colliding with other SVs). As a result, 9 samples lack corresponding paired data due to these collision events and are therefore excluded from the analysis. Among the 470 valid samples, 188 lack valid eye movement data (EMD) due to equipment disconnection or discontinuous data acquisition. Consequently, only 282 samples retain analyzable eye movement data. Therefore, the analysis of eye movement metrics is conducted exclusively on these 282 samples, whereas all 470 valid samples are utilized for analyzing the remaining metrics or indicators.

\subsection{Grouping of participants by defensive driving capability}
In prior research, defensive driving capability has been difficult to quantify and is typically characterized indirectly through surrogates such as driving experience. To overcome this restraint, this study evaluates defensive driving capability based on the observable actions taken by participants. Thanks to the precise definition of the defensive driving interval in Section \ref{sec1}, we are able to accurately classify participant actions into three distinct categories: voluntary adjustments (actions occurring before the PRP), active defense (actions within the PRP-ARP interval), and passive collision avoidance (actions occurring after the ARP).

\begin{figure}[pos=t] 
      \centering
      \includegraphics[width=.5\linewidth]{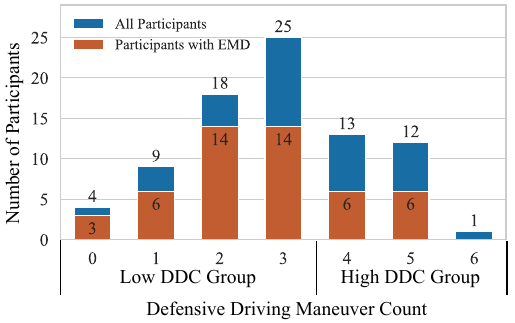}
      \caption{Distribution of participants by the number of successful defensive driving maneuvers performed across 6 driving samples.}
      \label{participant_group}
\end{figure}

To be more specific, within the PRP-ARP interval, we examine the collected data to determine whether participants exhibit any of the following defensive behaviors: 1) releasing the accelerator pedal, 2) pressing the brake pedal, or 3) executing a steering wheel turn > $10^{\circ}$. Notably, beyond the explicit risk-evasive actions (braking and steering), releasing the accelerator also contributes to deceleration and can be regarded as a preparation to brake. If any of these three conditions is met and a collision is successfully avoided, the participant is regarded to have executed a defensive driving maneuver in that sample. Given that each participant completes 6 driving samples (3 scenarios $\times$ 2 trials), we count the total number of samples in which defensive driving maneuver is performed. The distribution of participants according to these counts is depicted in Fig. \ref{participant_group}. As illustrated in the figure, the distribution approximates a normal distribution, suggesting that the data is representative and the measurement method possesses good reliability.

\begin{table}[pos=h]
\renewcommand{\arraystretch}{1.2}
\caption{Grouping of participants based on defensive driving capability.}
\label{tab4}
\begin{center}
\begin{tabular}{ m{0.2\linewidth} m{0.1\linewidth}<{\centering}  m{0.1\linewidth}<{\centering} m{0.01cm} m{0.1\linewidth}<{\centering} m{0.11\linewidth}<{\centering} }
\toprule
\multirow{2}{\linewidth}{} & \multicolumn{2}{c}{\textbf{Low DDC Group}} &  & \multicolumn{2}{c}{\textbf{High DDC Group}}   \\ \cline{2-3} \cline{5-6}
 & \textbf{P} & \textbf{N} & & \textbf{P} & \textbf{N} \\
\midrule
All valid data & 56 & 314 &  & 26 & 156 \\
Data with EMD & 37 & 210 & & 12 & 72 \\
\bottomrule
\end{tabular}
\begin{tablenotes}
\footnotesize
Note: P: number of participants, N: number of driving samples
\end{tablenotes}
\end{center}
\end{table}

Based on the results shown in Fig. \ref{participant_group}, participants who successfully execute defensive driving maneuvers three times or fewer ($\le$3) are assigned to the low defensive driving capability group (hereafter referred to as the low DDC group), whereas those with more than three successes ($>$3) are assigned to the high defensive driving capability group (hereafter referred to as the high DDC group). This grouping criterion is justified because membership in the high DDC group indicates that the participant successfully performs at least one defensive maneuver in the first trial (i.e., across the three scenarios). This suggests a pre-existing capability rather than one acquired solely through learning effects during the experiment. 
It is important to note that participants in the low DDC group may still exhibit defensive behaviors in certain driving samples. Conversely, those in the high DDC group may fail to exhibit such behaviors or even experience collisions in specific instances. Therefore, subsequent data analysis is conducted based on statistical significance rather than on individual performance.
Furthermore, through Fig. \ref{participant_group} we can find that the absence of eye movement data does not significantly alter the ratio of participants in the high and low DDC groups. The final composition of the groups and the distribution of driving samples are summarized in Table \ref{tab4}.

\subsection{Inter-group comparison results} \label{group_comparison}

After grouping the participants, differences between the high and low defensive driving capability groups are analyzed first, with the results presented in Table \ref{tab5} and \ref{tab6}.

As shown in Table \ref{tab5}, the Mann-Whitney U test is employed to analyze continuous metrics. Metrics with $p$<0.05 are considered statistically significant, whereas those showing no significant difference are shaded in gray. 
It is worth mentioning that since risk information transmission involves a process from risk identification to human action and finally to vehicle response, kinematic metrics may not respond rapidly enough to manifest changes within the short PRP-ARP interval. Therefore, the study scope for kinematic metrics in Table \ref{tab5} covers the PRP to scenario-end interval. In contrast, eye movement metrics correspond to the driver’s response during the first stage (risk identification) of information transfer process and can react rapidly. Thus, their study scope is confined to the PRP-ARP interval. Experimental results corroborate this rationale, as within the PRP-ARP interval, only the maximum longitudinal deceleration (max Dec) is observed to differ significantly between the two groups ($p$ < 0.001).
Additionally, the sample size for the risk indicator PET and the behavioral indicator TCT is reduced because some samples failed to meet the calculation requirements (e.g., no trajectory conflict between the ego vehicle and the interacting object, or failure to complete the driving task. See Section \ref{sec_data_analysis} for specific criteria).

\begin{figure}[pos=!b] 
      \centering
      \includegraphics[width=.55\linewidth]{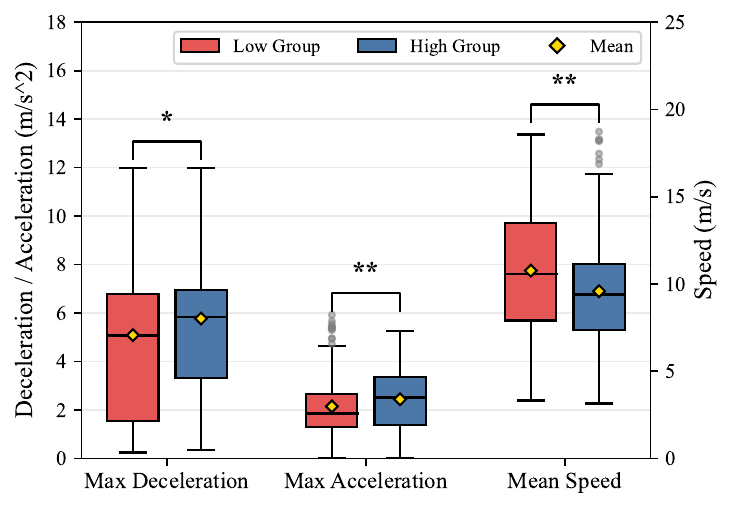}
      \caption{Comparison of longitudinal kinematic indicators between high and low defensive driving capability groups. *p<0.05, **p<0.01.}
      \label{kinematic_metrics}
\end{figure}

Significant differences are observed in both lateral and longitudinal kinematic metrics. Specifically, the high DDC group exhibits higher longitudinal acceleration and deceleration compared to the low DDC group, whereas their average longitudinal speed is significantly lower, as shown in Fig. \ref{kinematic_metrics}. Similarly, regarding lateral kinematic metrics, the high DDC group exhibits a larger mean absolute yaw angle. However, although the high DDC group have higher mean values for the remaining kinematic metrics, these differences are not statistically significant.

\begin{table}[pos=t]
\renewcommand{\arraystretch}{1.2}
\caption{Inter-group comparison results based on the Mann–Whitney U test.}
\label{tab5}
\begin{center}
\begin{tabular}{
  m{0.1\linewidth}
  m{0.1\linewidth}
  m{0.04\linewidth}
  m{0.03\linewidth}<{\centering}
  m{0.07\linewidth}<{\raggedleft}
  m{0.07\linewidth}<{\raggedleft}
  m{0.01cm}
  m{0.03\linewidth}<{\centering}
  m{0.07\linewidth}<{\raggedleft}
  m{0.07\linewidth}<{\raggedleft}
  m{0.06\linewidth}<{\centering}
  m{0.05\linewidth}<{\centering}
}
\toprule
\multirow{2}{\linewidth}{\textbf{Types}}
& \multirow{2}{\linewidth}{\textbf{Name}}
& \multirow{2}{\linewidth}{\textbf{Unit}}
& \multicolumn{3}{c}{\textbf{Low DDC Group}}
& & \multicolumn{3}{c}{\textbf{High DDC Group}}
& \multirow{2}{*}{\textbf{Z}}
& \multirow{2}{*}{$\boldsymbol{p}$\textbf{-value}} \\
\cline{4-6}\cline{8-10}
& & & \textbf{N} & \textbf{Mean} & \textbf{Median} & & \textbf{N} & \textbf{Mean} & \textbf{Median} & & \\
\midrule

\multirow{9}{\linewidth}{Kinematic Metrics (PRP$\rightarrow$END)}
& max Dec      & $m/s^2$ & 314 & 5.094 & 5.091 & & 156 & 5.772 & 5.842 & -2.181 & 0.0292 \\
& max Acc      & $m/s^2$ & 314 & 2.149 & 1.862 & & 156 & 2.440 & 2.504 & -2.580 & 0.0098 \\
& mean Speed   & $m/s$   & 314 & 10.730 & 10.528 & & 156 & 9.583 & 9.397 & 3.043 & 0.0023 \\

&\textcolor{gray!60}{max Speed}
& \textcolor{gray!60}{$m/s$}
& \textcolor{gray!60}{314}
& \textcolor{gray!60}{14.660}
& \textcolor{gray!60}{14.608}
& & \textcolor{gray!60}{156}
& \textcolor{gray!60}{14.670}
& \textcolor{gray!60}{14.712}
& \textcolor{gray!60}{-0.365}
& \textcolor{gray!60}{0.7154} \\

& mean Yaw      & $deg$   & 314 & 1.220 & 0.882 & & 156 & 1.358 & 1.092 & -1.962 & 0.0498 \\

&\textcolor{gray!60}{SD Yaw}
& \textcolor{gray!60}{$deg$}
& \textcolor{gray!60}{314}
& \textcolor{gray!60}{1.125}
& \textcolor{gray!60}{0.617}
& & \textcolor{gray!60}{156}
& \textcolor{gray!60}{1.318}
& \textcolor{gray!60}{0.930}
& \textcolor{gray!60}{-1.795}
& \textcolor{gray!60}{0.0727} \\

&\textcolor{gray!60}{max Yaw}
& \textcolor{gray!60}{$deg$}
& \textcolor{gray!60}{314}
& \textcolor{gray!60}{3.858}
& \textcolor{gray!60}{2.324}
& & \textcolor{gray!60}{156}
& \textcolor{gray!60}{4.510}
& \textcolor{gray!60}{3.331}
& \textcolor{gray!60}{-1.788}
& \textcolor{gray!60}{0.0739} \\

&\textcolor{gray!60}{max LD}
& \textcolor{gray!60}{$m$}
& \textcolor{gray!60}{314}
& \textcolor{gray!60}{1.300}
& \textcolor{gray!60}{0.980}
& & \textcolor{gray!60}{156}
& \textcolor{gray!60}{1.339}
& \textcolor{gray!60}{0.957}
& \textcolor{gray!60}{-0.225}
& \textcolor{gray!60}{0.8223} \\

&\textcolor{gray!60}{SDLD}
& \textcolor{gray!60}{$m$}
& \textcolor{gray!60}{314}
& \textcolor{gray!60}{0.417}
& \textcolor{gray!60}{0.334}
& & \textcolor{gray!60}{156}
& \textcolor{gray!60}{0.429}
& \textcolor{gray!60}{0.318}
& \textcolor{gray!60}{0.033}
& \textcolor{gray!60}{0.9738} \\
\hline

\multirow{10}{\linewidth}{Eye Movement Metrics (PRP$\rightarrow$ARP)}
& APD & $pixel$ & 210 & 48.210 & 47.980 & & 72 & 43.616 & 42.939 & 4.954 & 0.0000 \\
& MPD & $pixel$ & 210 & 50.816 & 50.069 & & 72 & 46.272 & 46.390 & 4.775 & 0.0000 \\
& AFD & $ms$    & 210 & 991.123 & 836.810 & & 72 & 943.170 & 752.107 & 2.266 & 0.0235 \\

&\textcolor{gray!60}{ASD}
& \textcolor{gray!60}{$ms$}
& \textcolor{gray!60}{210}
& \textcolor{gray!60}{41.418}
& \textcolor{gray!60}{27.261}
& & \textcolor{gray!60}{72}
& \textcolor{gray!60}{39.851}
& \textcolor{gray!60}{29.437}
& \textcolor{gray!60}{-0.829}
& \textcolor{gray!60}{0.4074} \\

& MFD & $ms$ & 210 & 3829.926 & 3579.000 & & 72 & 3451.757 & 2899.750 & 2.233 & 0.0256 \\

&\textcolor{gray!60}{MSD}
& \textcolor{gray!60}{$ms$}
& \textcolor{gray!60}{210}
& \textcolor{gray!60}{227.348}
& \textcolor{gray!60}{50.500}
& & \textcolor{gray!60}{72}
& \textcolor{gray!60}{240.458}
& \textcolor{gray!60}{54.000}
& \textcolor{gray!60}{-0.726}
& \textcolor{gray!60}{0.4676} \\

&\textcolor{gray!60}{MSA}
& \textcolor{gray!60}{$deg$}
& \textcolor{gray!60}{210}
& \textcolor{gray!60}{9.362}
& \textcolor{gray!60}{7.155}
& & \textcolor{gray!60}{72}
& \textcolor{gray!60}{9.654}
& \textcolor{gray!60}{8.459}
& \textcolor{gray!60}{-1.398}
& \textcolor{gray!60}{0.1625} \\

&\textcolor{gray!60}{SAS}
& \textcolor{gray!60}{$deg$}
& \textcolor{gray!60}{210}
& \textcolor{gray!60}{0.623}
& \textcolor{gray!60}{0.516}
& & \textcolor{gray!60}{72}
& \textcolor{gray!60}{0.664}
& \textcolor{gray!60}{0.641}
& \textcolor{gray!60}{-1.664}
& \textcolor{gray!60}{0.0963} \\

& FF  & $Hz$ & 210 & 1.313 & 1.180 & & 72 & 1.613 & 1.357 & -2.154 & 0.0313 \\

&\textcolor{gray!60}{SF}
& \textcolor{gray!60}{$Hz$}
& \textcolor{gray!60}{210}
& \textcolor{gray!60}{1.136}
& \textcolor{gray!60}{1.038}
& & \textcolor{gray!60}{72}
& \textcolor{gray!60}{1.308}
& \textcolor{gray!60}{1.089}
& \textcolor{gray!60}{-0.595}
& \textcolor{gray!60}{0.5522} \\
\hline

\multirow{3}{\linewidth}{Risk Indicators}
& min TTC  & $s$ & 314 & 1.789 & 1.405 & & 156 & 2.274 & 1.796 & -2.811 & 0.0045 \\

&\textcolor{gray!60}{max DNDA}
& \textcolor{gray!60}{/}
& \textcolor{gray!60}{314}
& \textcolor{gray!60}{0.631}
& \textcolor{gray!60}{0.567}
& & \textcolor{gray!60}{156}
& \textcolor{gray!60}{0.584}
& \textcolor{gray!60}{0.531}
& \textcolor{gray!60}{1.599}
& \textcolor{gray!60}{0.1088} \\

& PET & $s$ & 177 & 1.658 & 0.826 & & 93 & 2.891 & 2.734 & -4.464 & 0.0000 \\
\hline

\multirow{2}{\linewidth}{Behavioral Indicators}
& DistAct & $m$ & 314 & 44.205 & 40.916 & & 156 & 51.208 & 50.823 & -3.725 & 0.0002 \\

&\textcolor{gray!60}{TCT}
& \textcolor{gray!60}{$s$}
& \textcolor{gray!60}{207}
& \textcolor{gray!60}{21.429}
& \textcolor{gray!60}{20.924}
& & \textcolor{gray!60}{123}
& \textcolor{gray!60}{22.378}
& \textcolor{gray!60}{21.561}
& \textcolor{gray!60}{-1.671}
& \textcolor{gray!60}{0.0948} \\

\bottomrule
\end{tabular}
\end{center}
\end{table}

\begin{figure}[pos=t]
\centering

\subfigure[Static blind spot scenario - Low group]{
  \label{fig:static_low}
  \includegraphics[width=0.47\linewidth]{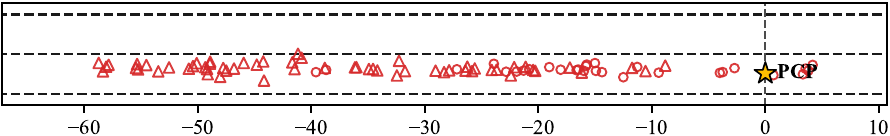}
}
\hspace{0.01\linewidth}
\subfigure[Static blind spot scenario - High group]{
  \label{fig:static_high}
  \includegraphics[width=0.47\linewidth]{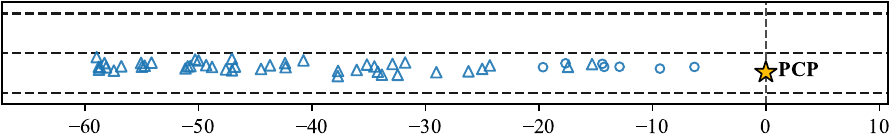}
}
\subfigure[Dynamic blind spot scenario - Low group]{
  \label{fig:dynamic_low}
  \includegraphics[width=0.47\linewidth]{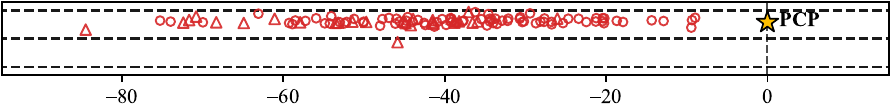}
}
\hspace{0.01\linewidth}
\subfigure[Dynamic blind spot scenario - High group]{
  \label{fig:dynamic_high}
  \includegraphics[width=0.47\linewidth]{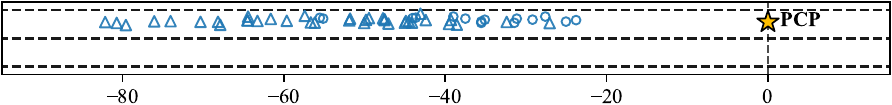}
}
\subfigure[Predictable motion uncertainty - Low group]{
  \label{fig:pred_low}
  \includegraphics[width=0.47\linewidth]{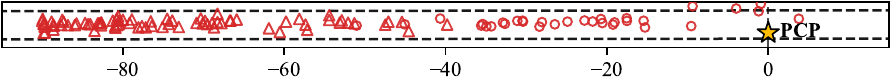}
}
\hspace{0.01\linewidth}
\subfigure[Predictable motion uncertainty - High group]{
  \label{fig:pred_high}
  \includegraphics[width=0.47\linewidth]{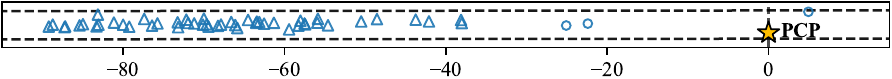}
}

\caption{Illustration of the spatial distribution of first action initiation points. $\triangle$ denotes actions taken within the PRP-ARP interval, whereas $\bigcirc$ denotes actions taken after the ARP. }
\label{first_action}
\end{figure}

As for eye movement metrics, significant differences are observed in the following metrics: the high DDC group exhibits smaller average and maximum pupil diameters, shorter average and maximum fixation durations, and a higher fixation frequency. In contrast, no significant differences are found in saccade-related indicators. With respect to the three risk indicators, both min TTC and PET show significant differences between the groups, with the high DDC group exhibiting higher mean values for both. Finally, regarding the behavioral indicators, although no significant difference is found in task completion time, the high DDC group demonstrates a greater distance to the PCP at the moment of first action initiation. Fig. \ref{first_action} further illustrates the spatial distribution of first action initiation points for both groups across the three scenarios, with the yellow stars representing the PCP for each scenario as defined in Table \ref{tab2}. As depicted in Fig. \ref{first_action}, participants in the high DDC group initiate actions at a greater distance from the PCP, resulting in a significantly larger DistAct. Furthermore, a higher proportion of the high DDC group executes actions within the PRP-ARP interval, whereas a considerable portion of the low DDC group delays their responses until after the ARP.

For binary indicators, the Chi-square test is applied, with the results shown in Table \ref{tab6}. Although a significant difference exists between the groups regarding collision occurrence, this indicator is not meaningful for inter-group comparison since our grouping criteria already incorporate collision avoidance. As for task completion, no significant difference exists between the two groups once collision samples are excluded.

\begin{table}[pos=t]
\renewcommand{\arraystretch}{1.2}
\caption{Inter-group comparison results for binary indicators based on the Chi-square test.}
\label{tab6}
\begin{center}
\begin{tabular}{ m{0.1\linewidth} m{0.1\linewidth}<{\centering}  m{0.1\linewidth}<{\centering} m{0.001\linewidth} m{0.1\linewidth}<{\centering} m{0.1\linewidth}<{\centering} m{0.09\linewidth}<{\centering} m{0.09\linewidth}<{\centering} } 
\toprule
\multirow{2}{*}{\textbf{Indicator}} & \multicolumn{2}{c}{\textbf{Low DDC Group}} & & \multicolumn{2}{c}{\textbf{High DDC Group}} & \multirow{2}{*}{$\boldsymbol{\chi^2}$} & \multirow{2}{*}{$\boldsymbol{p}$\textbf{-value}}    \\ \cline{2-3} \cline{5-6}
& True & False & & True & False & &  \\
\midrule

Collision & 68 & 246 & & 15 & 141 & 10.3912 & 0.0013 \\
\textcolor{gray!60}{TC} 
&\textcolor{gray!60}{207} 
&\textcolor{gray!60}{39}  
& &\textcolor{gray!60}{123} 
&\textcolor{gray!60}{18} 
&\textcolor{gray!60}{0.6804} 
&\textcolor{gray!60}{0.4095} \\

\bottomrule
\end{tabular}
\end{center}
\end{table}

\subsection{Inter-trial comparison results}

During the experiment, each participant completes two trials for the same scenario. Since they experience the full scenario in the first trial (T1), participants can learn about potential risks and how these risks evolve. By the second trial (T2), participants are considered to possess defensive driving capabilities for that specific scenario (meaning they can anticipate and accurately identify potential risks). Consequently, the differences between the two trials offer valuable insights into defensive driving. As a result, inter-trial comparison is conducted, with the results shown in Table \ref{tab7} and \ref{tab8}.

Since the data from T1 and T2 for each participant constitute paired samples, we employ the Wilcoxon Signed-rank test for continuous indicators. As shown in Table \ref{tab7}, the results include the number of paired samples per DDC group, the mean difference between T1 and T2 (T2 mean - T1 mean), the effect size, and the $p$-value. Similar to Table \ref{tab5}, data with $p$ > 0.05 are shaded in gray and are considered non-significant. 

\begin{table}[pos=t]
\renewcommand{\arraystretch}{1.2}
\caption{Inter-trial comparison results based on the Wilcoxon Signed-rank test.}
\label{tab7}
\begin{center}
\begin{tabular}{
  m{0.1\linewidth}
  m{0.1\linewidth}
  m{0.04\linewidth}
  m{0.04\linewidth}<{\centering}
  m{0.07\linewidth}<{\raggedleft}
  m{0.06\linewidth}<{\raggedleft}
  m{0.06\linewidth}<{\centering}
  m{0.01cm}
  m{0.04\linewidth}<{\raggedleft}
  m{0.07\linewidth}<{\raggedleft}
  m{0.06\linewidth}<{\centering}
  m{0.06\linewidth}<{\centering}
}
\toprule
\multirow{2}{\linewidth}{\textbf{Types}}
& \multirow{2}{\linewidth}{\textbf{Name}}
& \multirow{2}{\linewidth}{\textbf{Unit}}
& \multicolumn{4}{c}{\textbf{Low DDC Group}}
& & \multicolumn{4}{c}{\textbf{High DDC Group}} \\
\cline{4-7}\cline{9-12}
& & & \textbf{N}(pairs) & \textbf{Diff.} & \textbf{Effect} & $\boldsymbol{p}$\textbf{-value}
& & \textbf{N}(pairs) & \textbf{Diff.} & \textbf{Effect} & $\boldsymbol{p}$\textbf{-value} \\
\midrule

\multirow{9}{\linewidth}{Kinematic Metrics (PRP$\rightarrow$END)}

& max Dec & $m/s^2$ & 
\textcolor{gray!60}{157} &
\textcolor{gray!60}{-0.090} &
\textcolor{gray!60}{-0.019} &
\textcolor{gray!60}{0.8368} & &
\textcolor{gray!60}{78} &
\textcolor{gray!60}{-0.075} &
\textcolor{gray!60}{-0.056} &
\textcolor{gray!60}{0.6702} \\

& max Acc & $m/s^2$ & 
\textcolor{gray!60}{157} &
\textcolor{gray!60}{0.010} &
\textcolor{gray!60}{0.012} &
\textcolor{gray!60}{0.8947} & &
\textcolor{gray!60}{78} &
\textcolor{gray!60}{-0.031} &
\textcolor{gray!60}{-0.047} &
\textcolor{gray!60}{0.7180} \\

& mean Speed & $m/s$ & 157 & -0.899 & -0.184 & 0.0453 & &
\textcolor{gray!60}{78} &
\textcolor{gray!60}{-0.206} &
\textcolor{gray!60}{0.011} &
\textcolor{gray!60}{0.9305} \\

& max Speed & $m/s$ & 
\textcolor{gray!60}{157} &
\textcolor{gray!60}{-0.528} &
\textcolor{gray!60}{-0.167} &
\textcolor{gray!60}{0.0690} & &
\textcolor{gray!60}{78} &
\textcolor{gray!60}{-0.342} &
\textcolor{gray!60}{-0.214} &
\textcolor{gray!60}{0.1008} \\

& mean Yaw & $deg$ & 157 & 0.179 & 0.185 & 0.0441 & & 
\textcolor{gray!60}{78} &
\textcolor{gray!60}{-0.012} &
\textcolor{gray!60}{0.083} &
\textcolor{gray!60}{0.5254} \\

& SD Yaw & $deg$ & 157 & 0.233 & 0.242 & 0.0086 & & 
\textcolor{gray!60}{78} &
\textcolor{gray!60}{-0.044} &
\textcolor{gray!60}{0.008} &
\textcolor{gray!60}{0.9504} \\

& max Yaw & $deg$ & 157 & 0.756 & 0.242 & 0.0087 & & 
\textcolor{gray!60}{78} &
\textcolor{gray!60}{-0.037} &
\textcolor{gray!60}{0.011} &
\textcolor{gray!60}{0.9305} \\

& max LD & $m$ & 157 & 0.156 & 0.183 & 0.0472 & &
\textcolor{gray!60}{78} &
\textcolor{gray!60}{0.291} &
\textcolor{gray!60}{0.115} &
\textcolor{gray!60}{0.3793} \\

& SDLD & $m$ & 
\textcolor{gray!60}{157} &
\textcolor{gray!60}{0.036} &
\textcolor{gray!60}{0.163} &
\textcolor{gray!60}{0.0757} & &
\textcolor{gray!60}{78} &
\textcolor{gray!60}{0.089} &
\textcolor{gray!60}{0.091} &
\textcolor{gray!60}{0.4872} \\

\hline

\multirow{10}{\linewidth}{Eye Movement Metrics (PRP$\rightarrow$ARP)}

& APD & $pixel$ & 105 & 0.638 & 0.310 & 0.0059 & & 
\textcolor{gray!60}{36} &
\textcolor{gray!60}{0.130} &
\textcolor{gray!60}{0.252} &
\textcolor{gray!60}{0.1920} \\

& MPD & $pixel$ & 
\textcolor{gray!60}{105} &
\textcolor{gray!60}{0.416} &
\textcolor{gray!60}{0.164} &
\textcolor{gray!60}{0.1445} & &
\textcolor{gray!60}{36} &
\textcolor{gray!60}{-0.069} &
\textcolor{gray!60}{0.177} &
\textcolor{gray!60}{0.3624} \\

& AFD & $ms$ & 105 & 117.714 & 0.236 & 0.0358 & & 
\textcolor{gray!60}{36} &
\textcolor{gray!60}{218.897} &
\textcolor{gray!60}{0.352} &
\textcolor{gray!60}{0.0700} \\

& ASD & $ms$ & 105 & -7.462 & -0.346 & 0.0023 & & 
\textcolor{gray!60}{36} &
\textcolor{gray!60}{-4.629} &
\textcolor{gray!60}{-0.129} &
\textcolor{gray!60}{0.5207} \\

& MFD & $ms$ & 
\textcolor{gray!60}{105} &
\textcolor{gray!60}{161.710} &
\textcolor{gray!60}{0.100} &
\textcolor{gray!60}{0.3716} & &
\textcolor{gray!60}{36} &
\textcolor{gray!60}{550.181} &
\textcolor{gray!60}{0.252} &
\textcolor{gray!60}{0.1920} \\

& MSD & $ms$ & 105 & -78.267 & -0.414 & 0.0004 & & 
\textcolor{gray!60}{36} &
\textcolor{gray!60}{-33.583} &
\textcolor{gray!60}{-0.054} &
\textcolor{gray!60}{0.7806} \\

& MSA & $deg$ & 105 & -2.478 & -0.251 & 0.0256 & & 
\textcolor{gray!60}{36} &
\textcolor{gray!60}{0.942} &
\textcolor{gray!60}{-0.015} &
\textcolor{gray!60}{0.9443} \\

& SAS & $deg$ & 105 & -0.173 & -0.357 & 0.0015 & & 
\textcolor{gray!60}{36} &
\textcolor{gray!60}{0.009} &
\textcolor{gray!60}{-0.057} &
\textcolor{gray!60}{0.7740} \\

& FF & $Hz$ & 105 & -0.176 & -0.270 & 0.0163 & & 
\textcolor{gray!60}{36} &
\textcolor{gray!60}{-0.006} &
\textcolor{gray!60}{-0.018} &
\textcolor{gray!60}{0.9320} \\

& SF & $Hz$ & 105 & -0.215 & -0.400 & 0.0004 & & 
\textcolor{gray!60}{36} &
\textcolor{gray!60}{-0.000} &
\textcolor{gray!60}{-0.045} &
\textcolor{gray!60}{0.8220} \\

\hline

\multirow{3}{\linewidth}{Risk Indicators}

& min TTC & $s$ & 157 & 0.336 & 0.239 & 0.0195 & & 
\textcolor{gray!60}{78} &
\textcolor{gray!60}{0.239} &
\textcolor{gray!60}{0.124} &
\textcolor{gray!60}{0.3653} \\

& max DNDA & / & 157 & -0.072 & -0.265 & 0.0058 & &
\textcolor{gray!60}{78} &
\textcolor{gray!60}{-0.039} &
\textcolor{gray!60}{-0.118} &
\textcolor{gray!60}{0.3660} \\

& PET & $s$ & 70 & 0.795 & 0.403 & 0.0066 & & 
\textcolor{gray!60}{39} &
\textcolor{gray!60}{0.117} &
\textcolor{gray!60}{0.018} &
\textcolor{gray!60}{0.9285} \\
\hline

\multirow{2}{\linewidth}{Behavioral Indicators}

& DistAct & $m$ & 
\textcolor{gray!60}{157} &
\textcolor{gray!60}{1.361} &
\textcolor{gray!60}{0.067} &
\textcolor{gray!60}{0.4665} & &
\textcolor{gray!60}{78} &
\textcolor{gray!60}{1.078} &
\textcolor{gray!60}{0.093} &
\textcolor{gray!60}{0.4748} \\

& TCT & $s$ & 
\textcolor{gray!60}{80} &
\textcolor{gray!60}{-0.081} &
\textcolor{gray!60}{-0.065} &
\textcolor{gray!60}{0.6167} & &
52 & -0.894 & -0.416 & 0.0097 \\

\bottomrule
\end{tabular}
\end{center}
\end{table}

Evidently, the low DDC group exhibits significant inter-trial differences in most metrics, whereas the high DDC group shows almost no significant inter-trial differences across indicators. Specifically, regarding kinematic metrics, participants in the low DDC group demonstrate a lower average speed, larger absolute yaw angles (all three related metrics yield $p$<0.05), and a greater maximum lateral displacement in T2 compared to T1. For eye movement metrics, besides a subtle increase in average pupil diameter, the average fixation duration shows a significant increase in T2. Conversely, the mean and maximum saccade durations, along with the maximum saccade angle and its standard deviation, decrease. Simultaneously, both the saccade frequency and fixation frequency decline in T2. 
When it comes to risk indicators, all three indicators exhibit significant differences between T1 and T2. Both min TTC and PET increase in T2, whereas max DNDA decreases. Notably, based on the relationship between indicator values and scenario risk levels (scenario risk decreases as TTC and PET increase, whereas it increases as DNDA increases), these changes all signify a reduction in scenario risk in T2. Finally, the low DDC group shows no significant inter-trial differences in behavioral indicators. Interestingly, although no significant differences emerge in other metrics, the high DDC group achieves a significantly lower task completion time in T2 compared to T1.

Inter-trial comparison results for binary indicators based on the McNemar’s test is shown in Table \ref{tab8}. For the low DDC group, both binary indicators show significant differences between T2 and T1. Specifically, 35 paired samples that collide in T1 avoid collisions in T2, and 7 paired samples that fail the task in T1 complete it in T2. Although some samples perform worse in T2, statistical analysis indicates that the collision rate decreases and the task completion rate increases in T2. In contrast, no significant inter-trial differences are observed in either binary indicator for the high DDC group.

\begin{table}[pos=t]
\renewcommand{\arraystretch}{1.2}
\caption{Inter-trial comparison results for binary indicators based on the McNemar’s test.}
\label{tab8}
\begin{center}
\begin{tabular}{ m{0.07\linewidth} m{0.13\linewidth} m{0.05\linewidth}<{\centering} m{0.06\linewidth}<{\centering}  m{0.05\linewidth}<{\centering} m{0.05\linewidth}<{\centering} m{0.09\linewidth}<{\centering} m{0.09\linewidth}<{\centering} } 
\toprule
\multirow{2}{*}{\textbf{Group}} & \multirow{2}{*}{\textbf{Indicator}} & \multirow{2}{*}{\textbf{N}(p)} & \multirow{2}{*}{\textbf{T1}} & \multicolumn{2}{c}{\textbf{T2}} & \multirow{2}{*}{$\boldsymbol{\chi^2}$} & \multirow{2}{*}{$\boldsymbol{p}$\textbf{-value}}    \\\cline{5-6}
& & & & True & False & &   \\
\midrule

\multirow{4}{*}{Low} & \multirow{2}{*}{Collision} & \multirow{2}{*}{157} & True & 10 & 35 & \multirow{2}{*}{9.1875} & \multirow{2}{*}{0.0024} \\ \cline{4-6}
 & & & False & 13 & 99 & & \\ \cline{2-8}

 & \multirow{2}{*}{TC} &  \multirow{2}{*}{99} & True & 80 & 0 & \multirow{2}{*}{5.1429} & \multirow{2}{*}{0.0233} \\ \cline{4-6}
 & & & False & 7 & 12 & & \\ \hline

\multirow{4}{*}{\textcolor{gray!60}{High}} & \multirow{2}{*}{\textcolor{gray!60}{Collision}} & \multirow{2}{*}{\textcolor{gray!60}{78}} & \textcolor{gray!60}{True} & \textcolor{gray!60}{0} & \textcolor{gray!60}{11} & \multirow{2}{*}{\textcolor{gray!60}{2.4}} & \multirow{2}{*}{\textcolor{gray!60}{0.1213}} \\ \cline{4-6}
 & & & \textcolor{gray!60}{False} & \textcolor{gray!60}{4} & \textcolor{gray!60}{63} & & \\ \cline{2-8}

 & \multirow{2}{*}{\textcolor{gray!60}{TC}} &  \multirow{2}{*}{\textcolor{gray!60}{63}} & \textcolor{gray!60}{True} & \textcolor{gray!60}{52} & \textcolor{gray!60}{4} & \multirow{2}{*}{\textcolor{gray!60}{0.1667}} & \multirow{2}{*}{\textcolor{gray!60}{0.6831}} \\ \cline{4-6}
 & & & \textcolor{gray!60}{False} & \textcolor{gray!60}{2} & \textcolor{gray!60}{5} & & \\ 

\bottomrule
\end{tabular}
\begin{tablenotes}
\footnotesize
Note: N(p) represents the number of pairs.
\end{tablenotes}
\end{center}
\end{table}

\section{Discussion}

\textbf{Answer for RQ1}: Drivers with defensive driving capabilities tend to execute more intense braking, acceleration and steering maneuvers.

\begin{figure}[pos=b]%
    \centering
    \subfigure[Low DDC group]{
        \label{maddpg}
        \includegraphics[width=.6\linewidth]{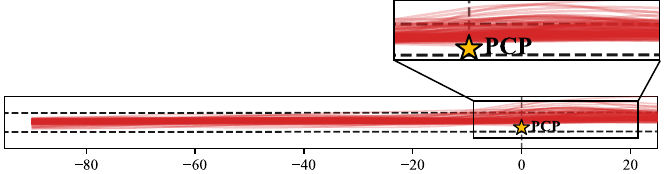}
        } 
    \subfigure[High DDC group]{
        \label{coma}
        \includegraphics[width=.6\linewidth]{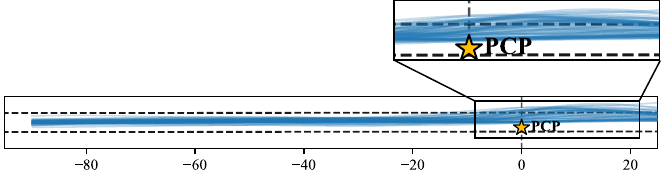}
        }
    \caption{Trajectory clusters of high and low DDC groups in predictable motion uncertainty scenario.}
    \label{predictable_traj}
\end{figure}

As described in Section \ref{group_comparison}, only maximum deceleration shows a significant inter-group difference within the PRP-ARP interval. This suggests that drivers with high defensive driving capability tend to brake more heavily upon detecting potential risks. Furthermore, as shown in Table \ref{tab5}, throughout the entire PRP to scenario-end interval, the high DDC group exhibits significantly higher maximum acceleration and deceleration, and lower mean longitudinal speed than the low DDC group. This implies that these drivers accelerate promptly after decelerating to mitigate risk, aiming to recover lost time, yet still end up with a lower average speed. 
Meanwhile, throughout the PRP to scenario-end interval, the high DDC group exhibits a significantly larger mean absolute yaw angle. This stems from drivers executing more frequent lateral evasive maneuvers followed by a return to the original lane, indicating that these drivers rely more on steering to mitigate risks. This tendency is well-reflected in predictable motion uncertainty scenarios. As illustrated in Fig. \ref{predictable_traj}, although drivers in both groups adopt lateral maneuvers to avoid the bicycle encroaching into their lane, a notably higher proportion of the high DDC group executes definitive steering actions. Furthermore, inter-trial comparison results in Table \ref{tab7} corroborate this phenomenon. After learning from T1, participants in the low DDC group exhibit more intense lateral behavior in T2. Concurrently, their mean maximum lateral displacement increases by nearly 0.16 m compared to T1.

\textbf{Answer for RQ2}: Defensive driving significantly enhances driving safety.

By analyzing accident reports, \cite{zhang2026intelligent} find that experienced drivers exhibit a significantly lower accident rate than novices, attributing this improvement to enhanced safety resulting from defensive driving experience accumulated through long-term practice. Extending this line of research, by utilizing both binary collision indicator and continuous risk indicators such as TTC, the current study investigates the impact of defensive driving on traffic safety directly from the perspective of defensive driving capability. As shown in Table \ref{tab5}, the high DDC group exhibits significantly higher min TTC and higher PET than the low DDC group, representing lower scenario risk and enhanced driving safety. Specifically, the high DDC group achieves an average min TTC approximately 0.5 seconds higher, and a PET nearly 1.3 seconds greater, than the low DDC group.

Inter-trial comparison results presented in Table \ref{tab7} and \ref{tab8} further demonstrate the positive contribution of defensive driving to driving safety. As evident in Table \ref{tab8}, after acquiring defensive driving capabilities, the collision rate for the low DDC group declines significantly in T2, providing direct evidence of improved safety. At the same time, all three risk indicators in Table \ref{tab7} shift toward lower risk levels in T2.

Furthermore, as described in Section \ref{sec_data_analysis}, pupil-related metric can reflect the driver's perception and response to risk or hazards. While \cite{qin2025eye} point out that vehicle occupants demonstrate significant pupil dilation in high hazard perception states, the current study observes significantly smaller average and maximum pupil diameters in the high DDC group compared to the low DDC group (Table \ref{tab5}). This indirectly confirms that defensive driving enhances driving safety, thereby preventing drivers from entering the high hazard perception states.

\textbf{Answer for RQ3}: The mechanism by which defensive driving enhances safety lies in advanced potential risk identification and early defensive actions.

Regarding potential risk identification, participants in the high DDC group exhibit shorter average and maximum fixation durations, as well as higher fixation frequencies (Table \ref{tab5}). This suggests that defensive drivers tend to switch rapidly among multiple fixation points so as to identify potential risks earlier. Moreover, as shown in Table \ref{tab7}, participants in the low DDC group reduce their saccade behavior (reflected in the average and maximum saccade duration, as well as the maximum and standard deviation of saccade angle) and fixation frequency in T2. This is because they acquire knowledge of the specific locations of potential risks in T1, they focus more intently on those areas to facilitate earlier identification of hazardous objects. In summary, both inter-group and inter-trial comparisons lead to a consistent conclusion: drivers possessing defensive driving capabilities engage in more proactive risk identification behaviors. 

\begin{figure}[pos=b]
\centering

\subfigure[Low DDC Group - T1]{
  \label{hm_lowT1}
  \includegraphics[width=0.46\linewidth]{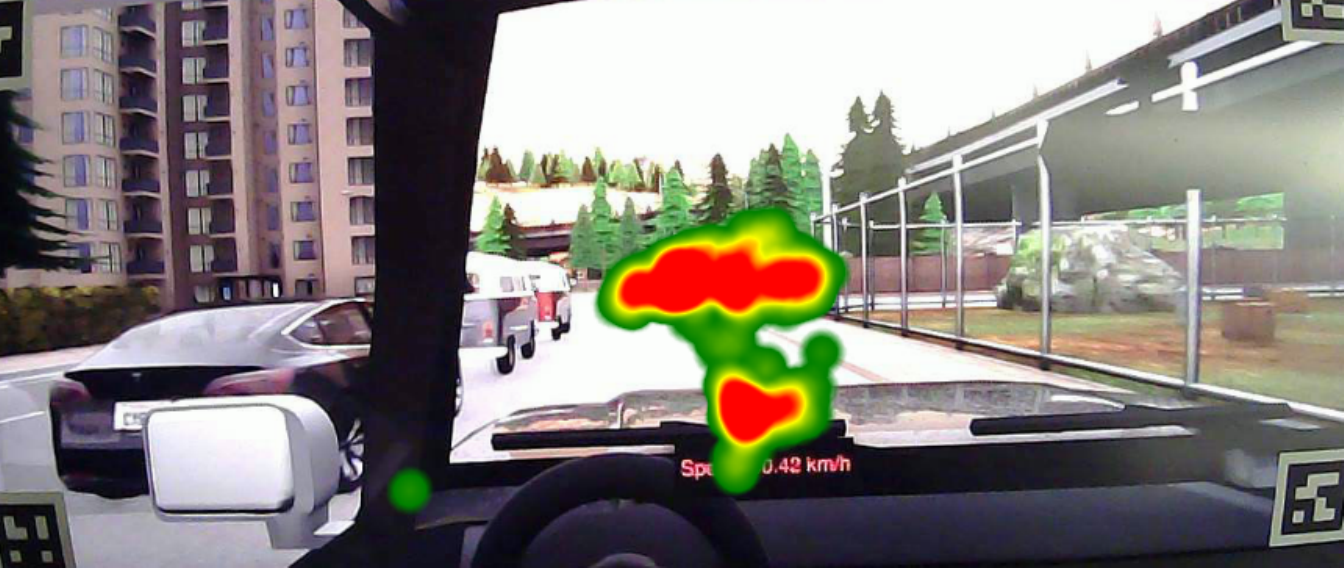}
}
\hspace{0.01\linewidth}
\subfigure[Low DDC Group - T2]{
  \label{hm_lowT2}
  \includegraphics[width=0.46\linewidth]{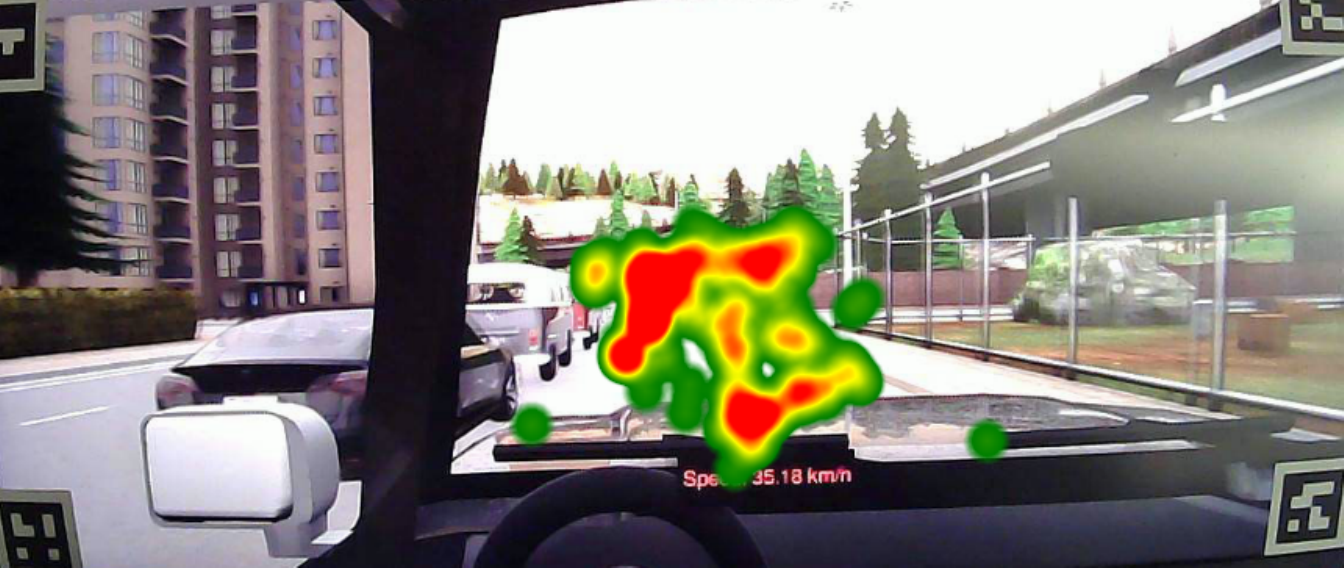}
}
\subfigure[High DDC Group - T1]{
  \label{hm_highT1}
  \includegraphics[width=0.46\linewidth]{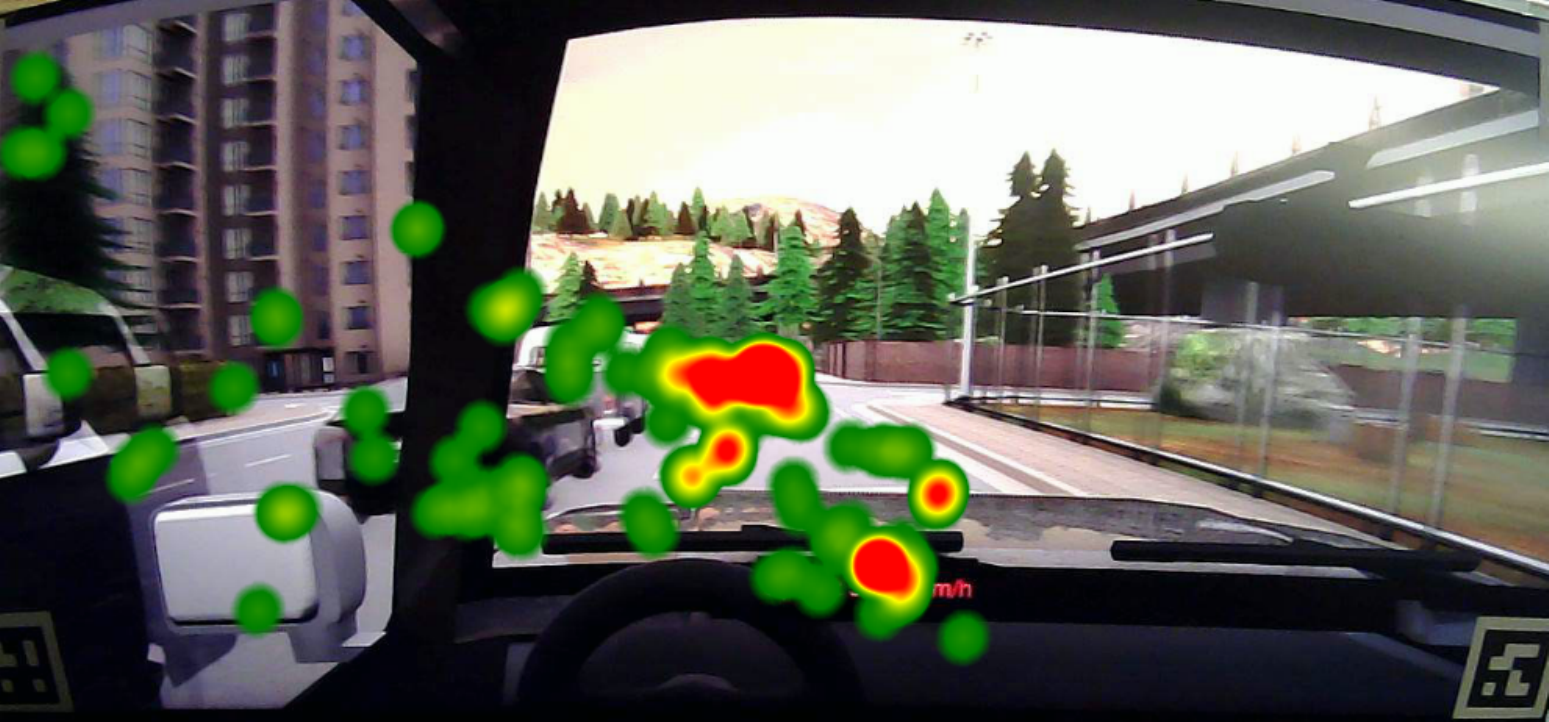}
}
\hspace{0.01\linewidth}
\subfigure[High DDC Group - T2]{
  \label{hm_highT2}
  \includegraphics[width=0.46\linewidth]{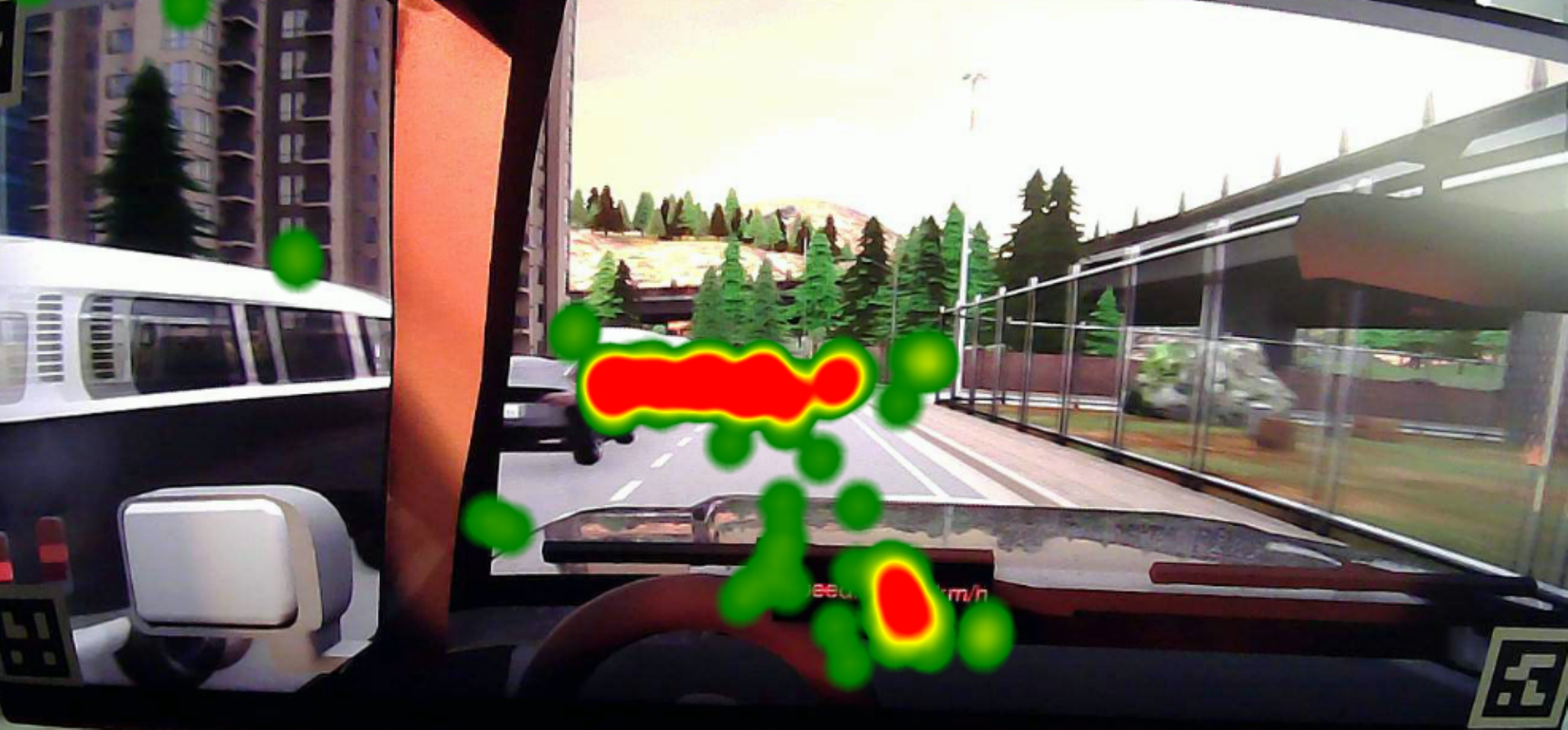}
}

\caption{Examples of eye movement heatmaps for one participant from the low and high defensive driving capability groups during T1 and T2 in the static blind spot scenario. }
\label{StaticBlind-Eyemovement}
\end{figure}

Fig. \ref{StaticBlind-Eyemovement} presents examples of eye movement heatmaps, corroborating the aforementioned discussion. As illustrated in Fig. \ref{hm_lowT1} and \ref{hm_highT1}, the low DDC participant in T1 concentrates fixations on the speedometer and the forward roadway, characterized by prolonged gazes. In contrast, the high capability participant in T1 exhibits significantly reduced fixation durations and distributes attention across multiple areas with brief glances, indicating active potential risk identification. Meanwhile, Fig. \ref{hm_lowT2} shows that the low group participant in T2, who has learned the risk location, demonstrates increased attention toward the area where the bicycle emerges. Furthermore, although no statistically significant inter-trial differences exist for the high DDC group's eye movement metrics, the specific example in Fig. \ref{hm_highT2} reveals that the participant allocates more attention to the area where the potential risk emerges in T2, confirming a learning effect. Moreover, these findings align with previous findings indicating that participants cast more glances toward the left than the right before entering the intersection \citep{ahlstrom2025driver}.

As for the initiation of maneuvers, while previous studies have found that experienced drivers are more likely to exhibit anticipatory driving behaviors (including pre-event actions and preparations) \citep{he2021anticipatorya}, the current study provides further empirical evidence for this phenomenon specifically from the perspective of defensive driving capability. As shown in Table \ref{tab5}, participants in the high DDC group conduct defensive/evasive actions at a significantly greater distance from the PCP, averaging approximately 7 meters earlier than the low DDC group (Table \ref{tab5}). This disparity is prominently illustrated in Fig. \ref{first_action}. Furthermore, in Fig. \ref{first_action}, a higher proportion of participants in the high DDC group initiate actions within the PRP–ARP interval. This suggests that, owing to earlier risk identification, these drivers are better able to execute proactive defensive actions rather than reactive evasive actions.

\textbf{Answer for RQ4}: Defensive driving capabilities can be acquired through training in scenarios with high potential risks.

Previous studies have demonstrated that training via driving simulators effectively enhances driving skills \citep{glassman2026comparable}, such as hazard awareness \citep{seibokaite2022improvement} and risk mitigation behaviors \citep{kuipers2026training}. Our research aligns with these findings and further reveals previously unreported phenomenon regarding defensive driving. In the inter-trial comparisons (Table \ref{tab7} and \ref{tab8}), the high DDC group exhibits no significant inter-trial differences in behavioral indicators or safety performance. This suggests that training offers limited improvement for drivers who already possess defensive skills. In contrast, the low DDC group demonstrates significant improvements across various indicators, accompanied by a notable enhancement in safety. This implies that for drivers lacking such skills, the simplest and most effective method of improvement involves exposing them to scenarios with high potential risks. By learning the locations and evolutionary patterns of these risks, drivers can anticipate and identify hazards earlier in similar future scenarios, thereby initiating defensive actions proactively.

\textbf{Additional discussion on driving efficiency}

Driving efficiency stands as the primary concern aside from safety itself \citep{shi2024deepad, tian2025framework}. A major concern regarding defensive driving is that it may introduce overly conservative maneuvers, thereby wasting time and reducing efficiency. This concern is partially substantiated by the current study: as shown in Table \ref{tab5}, the high DDC group exhibits a lower mean longitudinal speed, and although not statistically significant, their average task completion time is also higher than that of the low DDC group.

However, the results also reveal the positive impact of defensive driving on efficiency. In Table \ref{tab8}, low capability participants who acquire defensive skills after T1 demonstrate a significant increase in task completion rates in T2. Concurrently, in Table \ref{tab7}, the only metric showing a significant inter-trial difference for the high DDC group is the reduction in task completion time. This suggests that while further training yields limited safety improvements for already skilled drivers, it enables them to execute more optimal defensive maneuvers, thereby enhancing driving efficiency. Therefore, defensive driving can be regarded as a strategy that effectively balances driving safety with driving efficiency.

\textbf{Limitations and future work}

The primary limitation of this study lies in participant homogeneity. Although we recruited drivers as diverse as possible, the sample remains predominantly composed of university students. Additionally, due to the inevitable lack of realism inherent in driving simulators, behavioral data may differ from real-world conditions to some extent. Therefore, future real-world experiments involving a broader demographic of drivers (particularly professional drivers such as ride-hailing drivers) are necessary to collect defensive driving data.

\section{Conclusion}

Regarding defensive driving, this paper first defines two key scenario risk points (namely the PRP and ARP) and then provides a systematic and formal definition of defensive driving. Based on these definitions, this study classifies high potential risk scenarios applicable to defensive driving into two main categories: existence uncertainty scenarios and motion uncertainty scenarios, and further designs three specific experimental scenarios with precision. After that, the human driving data is collected through a driving simulator experiment and then analyzed to investigate behavioral characteristics of defensive driving and its impact on safety. Experimental results indicate that drivers possessing defensive driving capabilities tend to execute more intense acceleration, deceleration, and steering maneuvers. Moreover, defensive driving significantly reduces driving risk and collision rates, thereby exerting a positive influence on driving safety. Furthermore, analyses of eye movement metrics and drivers' behavioral indicators elucidate the underlying mechanism by which defensive driving enhances safety: such drivers identify potential risks earlier and initiate risk mitigation actions in advance. Finally, inter-trial comparison demonstrates that defensive driving is a learnable driving skill.

The findings of this study have significant implications for driving safety. First, this research substantiates the necessity of defensive driving, as recommended by numerous organizations and institutions. Furthermore, it demonstrates an effective approach to enhancing defensive driving capability: exposing drivers to diverse potential risk scenarios to clarify the occurrence and evolutionary patterns of risks. These insights provide a basis for subsequent defensive driving training programs. Moreover, regarding the rapid advancement of autonomous driving technology, this study offers valuable insights for training algorithms. By integrating high potential risk scenarios \citep{yan2026neuralmetric} involving existence and motion uncertainty into training datasets, autonomous driving algorithms can acquire defensive driving skills. This enables them to anticipate hazards and mitigate risks proactively, thereby enhancing overall traffic safety and facilitating the practical deployment of autonomous driving technologies.


\section*{Acknowledgments}
This study was supported by the National Key R\&D Program of China under Grant 2024YFB2505705, and the National Natural Science Foundation of China under Grant 52572482 and 52232015.

\printcredits

\bibliographystyle{cas-model2-names}

\bibliography{citelist}


\end{document}